\pdfoutput=1

\documentclass[11pt]{article}

\usepackage[]{emnlp2021}

\usepackage{times}
\usepackage{latexsym}

\usepackage[T1]{fontenc}
\usepackage[utf8]{inputenc}

\usepackage{microtype}

\usepackage{multirow}
\usepackage{multicol}
\usepackage{adjustbox}
\usepackage{booktabs}
\usepackage{amssymb}
\usepackage{amsthm}
\usepackage{amsmath}

\title{Towards Automatic Bias Detection in Knowledge Graphs}

\author{Daphna Keidar$^{1}$, Mian Zhong$^{1}$ \thanks{The first two authors contributed equally.} , Ce Zhang$^{1}$, Yash Raj Shrestha$^{1}$, Bibek Paudel$^{2}$\\
  $^{1}$ ETH Z\"urich, Switzerland; $^{2}$ Stanford University, U.S.A.\\
  \texttt{\{dkeidar, mzhong, yshrestha\}@ethz.ch}\\ \texttt{ce.zhang@inf.ethz.ch, yshrestha@ethz.ch, bibekp@cs.stanford.edu}
  }

\begin{document}
\maketitle 

\begin{abstract}
With the recent surge in social applications relying on knowledge graphs, the need for techniques to ensure fairness in KG based methods is becoming increasingly evident. Previous works have demonstrated that KGs are prone to various social biases, and have proposed multiple methods for debiasing them.
However, in such studies, the focus has been on debiasing techniques, while the relations to be debiased are specified manually by the user. As manual specification is itself susceptible to human cognitive bias, there is a need for a system capable of quantifying and exposing biases, that can support more informed decisions on what to debias. 
To address this gap in the literature, we describe a framework for identifying biases present in knowledge graph embeddings, based on numerical bias metrics. We illustrate the framework with three different bias measures on the task of profession prediction, and it can be flexibly extended to further bias definitions and applications. The relations flagged as biased can then be handed to decision makers for judgement upon subsequent debiasing. 
\end{abstract}
\section{Introduction}
Knowledge graphs (KGs) update and represent world knowledge in a structured and scalable format. They are commonly embedded into lower dimensional representations, namely knowledge graph embeddings (KGEs), which have successfully been applied in diverse applications such as personalized recommendations~\citep{liu2019survey}, question answering ~\citep{huang2019knowledge}, and enhancement of language modeling~\citep{DBLP:conf/acl/ZhangHLJSL19, DBLP:conf/emnlp/PetersNLSJSS19, baumgartner2018aligning}.
Following the proliferation of social applications relying on KGEs, the issue of fairness in KG based methods is a growing concern.

Recent works show that KGEs are inclined to manifest bias, and propose methods for debiasing them~\citep{DBLP:conf/emnlp/FisherMPC20, arduini2020adversarial, bose2019compositional}. 
However, these works implicitly assume that the relations to be debiased are chosen by the practitioner without quantification (e.g. based on social preconception), which may result in a sub-optimal decision. 
Reaching an informed decision on 
what to debias thus poses a challenge, and there is currently no empirical, data-driven method for identifying biased relations in KGs. In lack of such a system, some potential biases may go unnoticed while others are exaggerated. 

In this paper we aim to fill this gap, and present a framework for numerically identifying biases in KGEs. 
Our goal is to facilitate decision making by providing a table of bias scores on KG relations, as well as to encourage exploratory research in comprehending the nature of KGE biases. Practically, we describe and implement the framework using three bias measures that we derive from bias definitions from the domain of machine learning fairness.\footnote{The code will be released at: \url{https://github.com/mianzg/kgbiasdetec}} The relations and their corresponding bias scores can then guide practitioners when deciding which relations to debias.

We experiment and evaluate our framework's feasibility by implementing it for three bias definitions and applying it to the two benchmark datasets FB15K-237 \citep{fb15k237} and Wikidata5m \citep{wikidata5m}. 
\section{Why Automatic Bias Detection?}

\subsection{Aptly Deciding what to Debias}
Before debiasing a KGE, the user must choose which relations to debias. However, there is currently no method for uncovering which relations are prone to bias in a KG, and in previous works this selection was done mostly manually. Such manual selection of the relations to debias by the user can in itself be biased; for instance, the bias in a KG may potentially be rooted in non obvious relations such as ZIP codes \citep{zip-code} or a person's given name which can go unnoticed. It is therefore imperative to measure biases across a broad set of possibly sensitive relations, in an extensive and empirical manner. 
\subsection{Identifying the Sources of Bias}
Biases in KGEs can arise from multiple sources, including the data collection process for the KG, the chosen ontology, or the embedding method \citep{female_popes}. To help better understand the sources of bias, our bias measurement framework can be used to measure bias in different embeddings of the same KG. By doing so, we can examine which biases are apparent or amplified in certain embedding approaches, analyse whether the embedding method affects bias, and infer which biases are inherent to the KG itself. The output of our framework opens the door to further studies comparing biases across embedding methods and KGs, and serves as a step towards uncovering the bias sources.

\subsection{Comparing Bias Types}
Moreover, the machine learning literature includes a range of bias and fairness definitions \citep{fairnessdefs}. By extracting bias scores that are based on diverse bias definitions, we can empirically compare and analyse the relationships and correlations among them.

\begin{table*}[hbt!]
\centering
\footnotesize
\resizebox{\linewidth}{!}{ \begin{tabular}{@{}lrrrrrrrrrrrrrrrr@{}}
    \toprule
    &\multicolumn{4}{c}{ Demographic Parity Distance $\uparrow$ } & & \multicolumn{4}{c}{ Predictive Parity Distance $\uparrow$} & &
    \multicolumn{4}{c}{Translational Likelihood $\uparrow$}\\
    \midrule
    & TransE &  ComplEx & DistMult & RotatE & & TransE & ComplEx & DistMult & RotatE & &TransE & ComplEx & DistMult & RotatE\\
    \midrule
    gender  & 0.127  & 0.177 & 0.191 & 0.182 && 0.037 & 0.008  & 0.012 & 0.01 & & \textbf{1.566} & 0.047 & \textbf{0.208} & 0.010\\
    languages   & 0.190  & 0.201 & 0.227 & 0.211 && 0.0 & 0.0    & 0.0 & 0.0 & & 0.824 & \textbf{0.293} & 0.077 & \textbf{0.012}\\
    nationality & \textbf{0.280}  & \textbf{0.361} & \textbf{0.289} & \textbf{0.277} && \textbf{0.267} & \textbf{0.296} & \textbf{0.215} & \textbf{0.368} & & 0.724 & 0.289 & 0.076 & 0.007\\
    \bottomrule 
\end{tabular}

}
\caption{\label{tab:DBD} Aggregated bias scores of three implemented measures when predicting profession in FB15K-237. The arrow indicates the direction of larger bias. We measure the bias scores with respect to gender, languages, and nationality across four different embedding methods. This table suggests an investigation into debiasing nationality, as it has the highest bias scores in all embeddings. 
}
\end{table*}

\section{Our Framework for KGE Bias Identification} 
In this section, we begin with an overview of three specific bias measures we employ, which are defined over the relations in a KG. We then provide an overall description of our pipeline. 

\subsection{Preliminaries: Bias Measures}
\label{sec:bias-measurements}
There is a multitude of definitions for fairness and bias in the machine learning literature \citep{fairnessdefs}, and impossibility theorems have shown that they cannot all be simultaneously achieved \citep{impossibility_machine_fairness, kleinberg2016inherent}. In our model, we implemented three different definitions of fairness, that we formally describe below. Our framework can be easily extended to additional fairness definitions.

The first two measures are Predictive and Demographic Parity \citep{Mitchell_2021, Predictive_parity}, both common fairness metrics, which rely on a classification task. They measure the bias of sensitive relations via classification on a target relation. In our experiments, the classifier is trained to predict the target relation ``profession'' in order to measure bias in other relations.

Predictive Parity focuses on the classifier's precision, whereas Demographic Parity is useful when the underlying ground truth data is biased. These metrics are not specific to KGs and we describe below how their definition is extended to the KG setting. The third measure, Translational Likelihood Bias (TLB), is specifically tailored towards KGEs \citep{fisher2020measuring_bias_in_KGEs}. It leverages the score function used in KGE training to update entity embeddings and compute bias.

Formally, a knowledge graph (KG) is a set of facts represented by triples of the form $(h,r,t) \in \mathcal{E} \times \mathcal{R} \times \mathcal{E}$, where $\mathcal{E}$ denotes the set of entities and $\mathcal{R}$ denotes the set of relations. Each triple $(h,r,t)$ has a head entity $h$, a relation $r$, and a tail entity $t$, represented by embedding vectors. As we are concerned with fairness, we focus on the sub-graph containing solely human entities $\mathcal{H}$, and their associated relations as $R_{\mathcal{H}}$. There may exist a set of sensitive relations $S\subset R_{\mathcal{H}}$ related to humans, towards which we want to detect any biases. For Predictive and Demographic Parity, we also assume a classification task and a classifier which takes entity embeddings as input, together with a relation and predicts the corresponding tails. Lastly, we refer $s$ as a possibly sensitive relation in the following definitions, and take it as binary for simplicity.
\paragraph{Demographic Parity}
A classifier satisfies demographic parity with respect to a sensitive relation $s$, if the classifier's predictions, denoted $\hat{y}$, are independent of $s$.
Namely, demographic parity holds if $\mathbb{P}[\hat{y} = a| s = 1 ] = \mathbb{P}[\hat{y} = a|s = 0 ]$ for all possible predictions $a$. We can then measure the demographic parity distance (DPD) as 
\begin{equation}
    \text{DPD}(s,a) = |P[ \hat{y}  = a| s=1]- P[\hat{y} = a|s = 0 ]|
    \label{eq:DPD1}
\end{equation}
We finally compute 

\begin{equation}
    \text{DPD}(s) = \sum_{a} \text{DPD}(s,a)
    \label{eq:DPD}
\end{equation}
In the case of profession prediction, $\hat{y}$ stands for the predicted profession, and $a$ stands for a possible profession. Intuitively, DPD measures how much the sensitive relation affects classification, and therefore depends on the data itself. For instance, let the sensitive relation $s$ represent ``isChristian''.
In this setting, if all photographers in the KG were Christian, an accurate profession classifier would have a high DPD with respect to being Christian, and a random one would have low DPD. Namely, DPD rewards a classifier that is agnostic to the sensitive relation, regardless of its performance. 
\paragraph{Predictive Parity} A classifier satisfies this definition with respect to a sensitive relation $s$ if its precision is independent of $s$. 
Given a predicted label $a$, the predictive parity distance (PPD) is then defined as 
\begin{equation}
\begin{split}
\text{PPD}(s,a) & = |P[\hat{y} = a|y=a, s=1] \\
 & - P[\hat{y}=a|y=a, s = 0]|
\end{split}
\end{equation}
We then compute 
\begin{equation}
    \text{PPD}(s) = \sum_{a} \text{PPD}(s,a)
\end{equation}
\paragraph{Translational Likelihood} 
KGEs are typically trained using a score function $\phi$ that is unique to the embedding method, and captures closeness between entities and relations in the embedding space. According to \citet{fisher2020measuring_bias_in_KGEs}, we can directly use such score functions to measure bias. Given a triple $(h, r, t)$, we obtain a translated triple $(h', r, t)$ by performing a one-step gradient descent to update the head entity embedding on a direction of sensitive relation $s$. For instance, if $s$ is gender, we can translate the head entity $h$ in the direction of the entity ``female'' to obtain a new $h'$. The translational likelihood bias is then calculated as the difference in scores between the original and translated triples on a target relation $r_i$ 
\begin{align}
  \text{TL}(s,t_{r_i}) = \phi(h', r_i, t_{r_i}) - \phi(h, r_i, t_{r_i})
\end{align}
Following the example, a positive bias value indicates that making $h$ more ``female'' results in a higher score for the tail occupation $t_{r_i}$, and a negative bias value indicates that the new entity $h'$ scores worse on $t_{r_i}$. A score closer to zero suggests a relatively fairer relation, i.e. less bias.

Compared with DPD and PPD which are based on a downstream classification task, TLB does not require an external task to compute bias as it is calculated directly from the KGE. 
\subsection{Score Aggregation}
While the scores in section \ref{sec:bias-measurements} can be used to calculate bias for a binary sensitive relation, in practice, a sensitive relation $s$ may have multiple tail values $t$. 
In general, given a possibly sensitive relation $s$, we are interested in a score function
$\text{bias}_{\text{agg}}(s):\mathcal{R_{\mathcal{H}}} \rightarrow \mathbb{R}$
that will summarise how much bias there is towards $s \in \mathcal{R_{\mathcal{H}}}$. 

To generalize $\text{bias}_{\text{agg}}$ to the non-binary case, consider a sensitive relation $s$ and let $t_1,...t_n$ be all the possible tails $s$ can have. We aggregate the score over $t_i$;
\begin{equation}
   \text{bias}_{\text{agg}}(s) = \frac{1}{n}\sum_{i=1}^n \text{bias}(s = t_i)
\end{equation}
where $\text{bias}(s=t_i)$ is the bias with respect to having tail $t_i$ with the relation $s$, 
and is calculated according to the measure of choice. 
\subsection{Bias Detection Framework}
The workflow of our framework allows the user to specify a pre-trained KGE and a set of bias measurements. In the case of DPD and PPD, the user should also specify a classification task, namely the target relation to be classified. For clarity, throughout this paper we will consider profession to be the classification target. The output of our model is a table containing bias scores for each specified relation and bias measurement. This provides users with a ranking of relations according to their bias scores, which can empirically inform decisions on which relations to debias. After observing the bias scores for each relation, the practitioner can choose which ones to debias by according to their domain knowledge. Furthermore, our tool offers a multi-bias perspective for comparing bias across different embeddings, and helps select the appropriate embeddings according to downstream applications. 
\begin{table*}[hbt!]
\resizebox{\textwidth}{!}{
\begin{tabular}{@{}llrrrrrrrrr@{}}
    \toprule
    &\multicolumn{4}{c}{ Demographic Parity Distance } & & \multicolumn{4}{c}{ Predictive Parity Distance }\\
    \midrule
    & TransE  &  ComplEx & RotatE & DistMult & & TransE  &  ComplEx & RotatE & DistMult\\
    \midrule
    country of citizenship & 0.53 & 0.55 & 0.54 & 0.52 && 0.07 & 0.08 & 0.09 & 0.07 \\
    given name & 0.59 & 0.63 & 0.51 & 0.60 && \textbf{0.1} & 0.11 & 0.11 & 0.09 \\
    place of birth  & 0.52 & 0.51 & 0.47 & 0.51 && 0.06 & 0.08 & 0.02 & 0.06 \\
    sport & 0.73 & 0.74 & 0.78 & 0.64 && 0.0 & 0.0 & 0.0 & 0.0\\
    languages spoken & 0.46 & 0.57 & 0.49 & 0.54 && 0.07 & 0.09 & \textbf{0.14} & 0.08\\
    position played on team / speciality & \textbf{1.21} & 
    \textbf{1.26} & \textbf{1.24} & \textbf{1.11} && 0.06 & \textbf{0.14} & \textbf{0.14} & \textbf{0.12}\\
    \bottomrule
\end{tabular} 

}
\caption{Bias scores for most the common relations in Wikidata5m under a profession prediction task. The relation ``position played on team/specialty'' has the highest Demographic Parity Distance bias by a margin, which can be explained by its direct relation to profession. We further note that bias patterns are similar across embeddings.
}
\label{tab:wiki5m-dpd}
\end{table*}

\section{Experiments}
We applied our framework with the bias measurements described in section \ref{sec:bias-measurements} to evaluate bias on two benchmark datasets; FB15k-237 ~\citep{fb15k237} and Wikidata5m \citep{wikidata5m}. Each of the datasets was trained on four KGE methods respectively; TransE, CompleEx, DistMult and RotatE. %
The embeddings for FB15K-237 were trained using the entire dataset, through pykeen's hyperparameter optimization pipeline. For Wikidata5m, we use pre-trained embeddings from GraphVite\footnote{the embeddings can be found at \url{https://graphvite.io/docs/latest/pretrained\_model.html}} \citep{zhu2019graphvite}. 
To measure DPD and PPD, a random forest classifier was trained on the task of profession prediction in both datasets. We attempt classification of the $5$ and $10$ most common professions in FB15K-237 and Wikidata5m respectively, and relabel the rest as ``OTHER''. 
A pre-processing step was applied to remove any tails that appeared less than $10$ times in the test set. 
\subsection{Results}
Table \ref{tab:DBD} compares all three bias measurements of the three most common relations in the FB15k-237 dataset. We observe that across all embeddings, the relation \textit{gender} has the lowest DPD bias, and \textit{nationality} the highest bias in both DPD and PPD, suggesting it may need debiasing. Moreover, no PPD bias is detected for \textit{languages}. The common patterns across embeddings might imply that the biases do not arise from the embedding methods, but are rather inherent to the data itself or to the classifier. TLB presents a more mixed picture, with the most biased relation varying between embeddings. Since TLB is calculated using the score function of the embedding model, it is likely to be more sensitive to the KGE method.

The aggregated DPD and PPD bias scores on Wikidata5m are shown in Table \ref{tab:wiki5m-dpd}. The relation portraying highest DPD on this dataset by a margin is \textit{position played on team/specialty}, followed by \textit{sport}. While our framework would mark these two relations as biased according to DPD, the practitioner might choose not to debias them, since they are related to a person's profession. Notably the PPD for these two relations is low, further illustrating the importance of offering a multi-bias perspective for a more robust bias evaluation. On the other hand, \textit{given name} scores relatively high on both DPD and PPD in most embeddings, and can be considered an unwanted bias by the practitioner, since a person's given name should normally not affect their occupation. Therefore, given these scores, one may choose to only debias the relation \textit{given name} in Wikidata5m. 

Lastly, we provide a qualitative example shown in Table \ref{tab:transE-nat}, presenting the disaggregated TLB bias scores with respect to \textit{nationality} in FB15k-237. We display the five professions with highest TLB with respect to England versus the United States, the two most common tails for the relation \textit{nationality}. At the fine-grained level, we notice a historical stereotype might remain, where England associates more with scientific occupations while the U.S. is biased towards entertainment careers. Moreover, the bias towards England appears lower, namely, the highest TLB bias towards England is significantly lower than the highest bias towards the U.S. The disagreggated tables presenting TLB bias for the other embedding methods and relations can be found in the Appendix \ref{sec:appendix3}.
\begin{table}[htbp!]
    \begin{center}
        \resizebox{\linewidth}{!}{ 
\begin{tabular}{@{}lrlr@{}}
    \toprule
     \multicolumn{2}{c}{ England } & \multicolumn{2}{c}{ U.S. } \\
     \midrule
     Mathematician & 0.0160 & Television director & 0.0250\\
     Biologist & 0.0158 & Television producer & 0.0227\\
     Football player & 0.0133 & Screenwriter & 0.0222\\
     Physician & 0.0105 & Radio personality & 0.0214\\
     Scientist & 0.0100 & Actor & 0.0207 \\
    \bottomrule
\end{tabular}
}
    \end{center}
        \caption{Professions with the highest Translational Likelihood Bias with respect to English versus U.S. nationalities in FB15K-237, using the TransE embedding. 
        }
        \label{tab:transE-nat} 
\end{table}

\section{Discussion}
In this paper, we proposed a novel framework to systematically identify, measure and inform biases in knowledge graph embeddings (KGE). The contribution of our model is to aid stakeholders and practitioners with a quantitative approach to identify biased relations in the KGE. Since biases are context- and culture-dependent, the final determination on what to debias may depend on the downstream task and is left to the practitioner. For example, one would want to remove gender biases from a question answering task about historical figures, while in medical related data, keeping gender information can be valuable for proper diagnosis.

Our implementation provides the user with bias scores rather than a binary decision. The choice of which relations to debias is then to be done by comparing the relative scores of relations in the KG, combined with domain knowledge. In future work, we would like to derive a threshold that can provide users with a binary score (biased/unbiased) for each relation, possibly through a statistical significance test. While a yes/no suggestion could save time and target a broader range of users, a careful analysis is required in order to define such a threshold without incurring further biases.

In summary, our paper presents a framework for quantifying bias in KGs, and by doing so identifies useful avenues for future research, and opens the possibility to compare various sources and definitions of bias. \citet{female_popes} raise the concern that debiasing is not a neutral task, but rather based on social norms and is at risk of becoming censorship. By presenting a numerical method for selecting which relations to debias, we aim to minimize these risks. We hope to have illuminated the importance of identifying bias, as a complimentary component to algorithms that mitigate it.

\bibliographystyle{acl_natbib}
\bibliography{custom}

\newpage

\appendix
\section{Appendix}
\label{sec:appendix}

\subsection{Profession Classifier}

We trained a random forest and an MLP classifier to predict the occupations on the KGEs. On random forest, we did hyperparameter search on maximal depths in $[3,4,5,6]$ and batch sizes in $[100,256,500]$. On FB15K-237, we chose the random forest classifier with maximal depth of $4$ and balanced class weights and a batch size of $256$ as our final model, as it had the best performance. On Wikidata5m we choose the MLP classifier. The accuracy and balanced accuracy on classifying each entity into $6$ and $11$ occupation classes on FB15K-237 and Wikidata5M respectively are presented in tables \ref{tab:rf} and \ref{tab:rf_wiki}.

\begin{table}[h!]
\centering
\resizebox{0.9\linewidth}{!}{
\begin{tabular}{c c c c c } 
 \hline
  & TransE &  ComplEx & DistMult & RotatE \\
 \hline
 accuracy  & 0.5 & 0.514 & 0.499 & 0.517 \\
 balanced accuracy & 0.329  & 0.34 & 0.33 & 0.356 \\
 \hline
\end{tabular}
}
\caption{\label{tab:rf} Performance of the random forest classifier on a 6 class classification task, predicting occupation on FB15K-237.
}
\end{table}

\begin{table}[h!]
\centering
\resizebox{0.9\linewidth}{!}{
\begin{tabular}{c c c c c } 
 \hline
  & TransE &  ComplEx & DistMult & RotatE \\
 \hline
 accuracy  & 0.7 & 0.67 & 0.68 &  0.63  \\
 balanced accuracy &  0.61 & 0.55 & 0.55 & 0.44 \\
 \hline
\end{tabular}
}
\caption{\label{tab:rf_wiki} Performance of the MLP classifier on a $11$ class classification task, predicting occupation on Wikidata5m.
}
\end{table}

\subsection{Knowledge Graph Embeddings}
For the purpose of this paper, we trained a range of knowledge graph embedding models on FB15K237. The hits@$k$ scores of the embeddings are listed in Table \ref{tbl:KGEscores} below. We trained the embeddings through the hyperparameter optimization pipeline of pykeen \citep{ali2020pykeen}, or by using the suggested parameters either from pykeen or openKE \citep{han2018openke}. 

\begin{table}[h!]
    \resizebox{0.9\linewidth}{!}{
\begin{tabular}{c c c c c c}
    \toprule
    & TransE & ConvE &  ComplEx & DistMult & RotatE \\
    \midrule
    hits@10 & 0.42 &0.308 & 0.183 &0.366 &0.446\\
    hits@3 & 0.271 &0.175& 0.183 &0.219 &0.289\\
    hits@1 & 0.094 &0.184 & 0.183 &0.118 &0.175\\
    \bottomrule
\end{tabular}
}
\caption{\label{tbl:KGEscores} Hit@k for the trained embeddings.}
\end{table}

\subsection{Translational Likelihood scores on FB15K-237}\label{sec:appendix3}
Below we present the disaggregated Translational Likelihood Bias (TLB) scores on FB15K-237, for the top three relations \textit{nationality}, \textit{language}, and \textit{gender}. 

\subsubsection{TransE}
In Table \ref{tab:transe-gender} and \ref{tab:transe-lang}, we provide results using TransE embeddings on FB15k-237 dataset.
\begin{table}[h]
        \resizebox{\linewidth}{!}{ \begin{tabular}{@{}lrlr@{}}
    \toprule
    \cmidrule{1-4}
     \multicolumn{2}{c}{ Male } & \multicolumn{2}{c}{ Female } \\
     \midrule
     Cinematographer & 0.0367 & Model & 0.0157 \\
     Farmer & 0.0365 & Pin-up model & 0.0131 \\
     Soldier & 0.0346 & Spokesperson & 0.0077 \\
     /m/0196pc & 0.0295 & VJ & 0.0044 \\
     Screenwriter & 0.0288 & Environmentalist & 0.0022 \\
    \bottomrule
\end{tabular}}
        \caption{\label{tab:transe-gender} Male v.s. Female.}
        \resizebox{\linewidth}{!}{ \begin{tabular}{@{}lrlr@{}}
    \toprule
    \cmidrule{1-4}
     \multicolumn{2}{c}{ English } & \multicolumn{2}{c}{ Hindi } \\
     \midrule
     Model & 0.0213 & Stunt performer & 0.0060\\
     Author & 0.0205 & Music Director & 0.0054\\
     Singer-songwriter & 0.0199 & Prime Minister of Canada & 0.0008\\
     Designer & 0.0199 & Politician & 0.0002 \\
     Spokesperson & 0.0189 & Storyboard artist &-0.0008\\
    \bottomrule
\end{tabular}}
        \caption{\label{tab:transe-lang} English language v.s. Hindi language.}
\end{table}

\subsubsection{ComplEx}
In Table \ref{tab:complex-gender}, \ref{tab:complex-lang} and \ref{tab:complex-nat}, we provide results using ComplEx embeddings on FB15k-237 dataset.
\begin{table}[h]
    \centering
        \resizebox{\linewidth}{!}{ \begin{tabular}{@{}lrlr@{}}
    \toprule
    \cmidrule{1-4}
     \multicolumn{2}{c}{ Male } & \multicolumn{2}{c}{ Female } \\
     \midrule
     Football Player & 0.0006 & Television Producer & 0.0015 \\
     Politician & 0.0003 & Comedian & 0.0014 \\
     Lawyer & 0.0002 & Prime Minister of Canada & 0.0014 \\
     Architect & 0.0002 & Television director & 0.0013 \\
     Mathematician & 0.0002 & Dub Actor & 0.0012 \\
    \bottomrule
\end{tabular}}
        \caption{\label{tab:complex-gender} Male v.s. Female.}

        \resizebox{\linewidth}{!}{ \begin{tabular}{@{}lrlr@{}}
    \toprule
    \cmidrule{1-4}
     \multicolumn{2}{c}{ English } & \multicolumn{2}{c}{ Hindi } \\
     \midrule
     Make-up artist & -0.0021 & Theatrical producer & 0.0046 \\
     Production sound mixer  & -0.0021 & Supermodel & 0.0046 \\
     Art Director & -0.0024 & Music video director & 0.0045 \\
     /m/089fss & -0.0025 & VJ & 0.0045 \\
     Football player & -0.0025 & Pin-up model & 0.0045 \\
    \bottomrule
\end{tabular}}
        \caption{\label{tab:complex-lang} English language v.s. Hindi language.}
\end{table}

\begin{table}[htb!]
        \resizebox{\linewidth}{!}{ \begin{tabular}{@{}lrlr@{}}
    \toprule
    \cmidrule{1-4}
     \multicolumn{2}{c}{ England } & \multicolumn{2}{c}{ U.S. } \\
     \midrule
     Architect & 0.0047 & Production sound mixer & -0.0012 \\
     Mathematician  & 0.0046 & Make-up artist & -0.0012 \\
     Scientist & 0.0046 & Voice Actor & -0.0017 \\
     Critic & 0.0046 & Art Director & -0.0017 \\
     Inventor & 0.0046 & Television producer & -0.0018 \\
    \bottomrule
\end{tabular}}
        \caption{\label{tab:complex-nat} England v.s. U.S..}
\end{table}

\subsubsection{DistMult}
In Table \ref{tab:distmult-gender}, \ref{tab:distmult-lang} and \ref{tab:distmult-nat}, we provide results using DistMult embeddings on FB15k-237 dataset.
\begin{table}[hbt!]
        \resizebox{\linewidth}{!}{ \begin{tabular}{@{}lrlr@{}}
    \toprule
    \cmidrule{1-4}
     \multicolumn{2}{c}{ Male } & \multicolumn{2}{c}{ Female } \\
     \midrule
       /m/0196pc & 0.0054 & Pin-up model & 0.0067\\
      Cinematographer & 0.005 &Model & 0.0059\\
      Soldier & 0.0049 & Supermodel & 0.0043\\
      /m/01c8w0 & 0.0049 & /m/064xm0 & 0.0026\\
      Mathematician  & 0.0046 & Prime Minister of Canada & 0.0024\\
    \bottomrule
\end{tabular}}
        \caption{\label{tab:distmult-gender} Male v.s. Female.}
        
        \resizebox{\linewidth}{!}{ \begin{tabular}{@{}lrlr@{}}
    \toprule
    \cmidrule{1-4}
     \multicolumn{2}{c}{ English } & \multicolumn{2}{c}{ Hindi } \\
     \midrule
     Author & 0.0021 & /m/028kk\textunderscore & 0.0005\\
     Artist & 0.002 & Costume designer & 0.0005\\
     Actor & 0.0019 & Audio engineer & 0.0002\\
     /m/0np9r & 0.0019 & Cinematographer & 0.0002\\
     Spokesperson & 0.0019 & Prime Minister of Canada & 0.0002\\
    \bottomrule
\end{tabular}}
        \caption{\label{tab:distmult-lang}  English language v.s. Hindi language.}
        \resizebox{\linewidth}{!}{ \begin{tabular}{@{}lrlr@{}}
    \toprule
    \cmidrule{1-4}
     \multicolumn{2}{c}{ England } & \multicolumn{2}{c}{ U.S. } \\
     \midrule
   Physician & 0.0013 & /m/0196pc & 0.0024\\
   Mathematician & 0.0013 & Screenwriter & 0.0022\\
   Scientist & 0.001 & Radio personality & 0.0022\\
   /m/0q04f & 0.0009 & /m/02krf9 & 0.0021\\
   Football player & 0.0008 & Animator & 0.002\\
   \bottomrule
\end{tabular}}
        \caption{\label{tab:distmult-nat} England v.s. U.S.}
\end{table}

\subsubsection{RotatE}
In Tables \ref{tab:rotate-gender}, \ref{tab:rotate-lang} and \ref{tab:rotate-nat}, we provide results using RotatE embeddings on FB15k-237 dataset.
\begin{table}[hbt!]
        \resizebox{\linewidth}{!}{ \begin{tabular}{@{}lrlr@{}}
    \toprule
    \cmidrule{1-4}
     \multicolumn{2}{c}{ Male } & \multicolumn{2}{c}{ Female } \\
     \midrule
       /m/0196pc  & 0.0002 &  Model  & 0.0002\\
     Cinematographer  & 0.0002 &  Pin-up model  & 0.0002 \\
     Inventor  & 0.0002 &  Supermodel  & 0.0002\\
     /m/01c8w0  & 0.0002 &  Spokesperson  & 0.0001\\
     Composer  & 0.0002 &  VJ  & 0.0001 \\
    \bottomrule
\end{tabular}}
        \caption{\label{tab:rotate-gender} Top 5 biased professions in terms of gender.}
        \resizebox{\linewidth}{!}{ \begin{tabular}{@{}lrlr@{}}
    \toprule
    \cmidrule{1-4}
     \multicolumn{2}{c}{ English } & \multicolumn{2}{c}{ Hindi } \\
     \midrule
     Theatrical producer   & 0.0002  &  /m/01tkqy   & 0.0 \\
     Spokesperson   & 0.0002  &  Politician   & 0.0 \\
     Author   & 0.0002  &  Prime Minister of Canada   & 0.0 \\
     Musician   & 0.0002  &  /m/028kk\textunderscore   & 0.0 \\
     Singer-songwriter   & 0.0002  &  Football player   & -0.0001 \\
    \bottomrule
\end{tabular}}
        \caption{\label{tab:rotate-lang} Top 5 biased professions: English language v.s. Hindi language.}
        \resizebox{\linewidth}{!}{ \begin{tabular}{@{}lrlr@{}}
    \toprule
    \cmidrule{1-4}
     \multicolumn{2}{c}{ England } & \multicolumn{2}{c}{ U.S. } \\
     \midrule
     Biologist & 0.0001 & Attorneys in the United States & 0.0002 \\
     Mathematician & 0.0001 &  /m/0196pc & 0.0002 \\
   Football player & 0.0001 & Music executive & 0.0002\\
    Physician & 0.0001 & Television producer & 0.0002\\
   /m/0q04f & 0.0001 & Businessperson & 0.0002 \\
    \bottomrule
\end{tabular}}
        \caption{\label{tab:rotate-nat} Top 5 biased professions.}
\end{table}

\end{document}